\pdfoutput=1
\documentclass{article}
\usepackage{iclr2016_conference,times}
\usepackage{hyperref}
\usepackage{url}
\usepackage{amsmath}
\usepackage{amssymb}
\usepackage{graphicx}
\usepackage{booktabs}
\usepackage{subcaption}
\usepackage{hyperref}
\usepackage{color}
\usepackage{multicol}

\hypersetup{
    bookmarks=true,         
    unicode=false,          
    pdftoolbar=true,        
    pdfmenubar=true,        
    pdffitwindow=false,     
    pdfstartview={FitH},    
    pdftitle={Reasoning about Entailment with Neural Attention},    
    pdfnewwindow=true,      
    colorlinks=true,        
    linkcolor=black,          
    citecolor=blue!50!black,    
    filecolor=black,      
    urlcolor=black           
}

\iclrfinalcopy

\title{Reasoning about Entailment with \\Neural Attention}

\author{Tim Rockt\"aschel \\
 University College London \\
 \texttt{t.rocktaschel@cs.ucl.ac.uk} \\
 \And
 Edward Grefenstette \& Karl Moritz Hermann \\
 Google DeepMind \\
 \texttt{\{etg,kmh\}@google.com} \\
 \And
 Tom\'a\v{s} Ko\v{c}isk\'y \& Phil Blunsom \\
 Google DeepMind \& University of Oxford \\
 \texttt{\{tkocisky,pblunsom\}@google.com} \\
}

\begin{document}

\maketitle

\begin{abstract}
While most approaches to automatically recognizing entailment relations have used classifiers employing hand engineered features derived from complex natural language processing pipelines, in practice their performance has been only slightly better than bag-of-word pair classifiers using only lexical similarity. The only attempt so far to build an end-to-end differentiable neural network for entailment failed to outperform such a simple similarity classifier.
  In this paper, we propose a neural model that reads two sentences to determine entailment using long short-term memory units.
  We extend this model with a word-by-word neural attention mechanism that encourages reasoning over entailments of pairs of words and phrases.
  Furthermore, we present a qualitative analysis of attention weights produced by this model, demonstrating such reasoning capabilities.
  On a large entailment dataset this model outperforms the previous best neural model and a classifier with engineered features by a substantial margin.
  It is the first generic end-to-end differentiable system that achieves state-of-the-art accuracy on a textual entailment dataset.
\end{abstract}

\section{Introduction}
The ability to determine the semantic relationship between two sentences is an integral part of machines that understand and reason with natural language.
Recognizing textual entailment (RTE) is the task of determining whether two natural language sentences are (i) contradicting each other, (ii) not related, or whether (iii) the first sentence (called \emph{premise}) entails the second sentence (called \emph{hypothesis}).
This task is important since many natural language processing (NLP) problems, such as information extraction, relation extraction, text summarization or machine translation, rely on it explicitly or implicitly and could benefit from more accurate RTE systems \citep{dagan2006pascal}.

State-of-the-art systems for RTE so far relied heavily on engineered NLP pipelines, extensive manual creation of features, as well as various external resources and specialized subcomponents such as negation detection \citep[e.g.][]{lai2014illinois, jimenez2014unal, zhao2014ecnu, beltagy2015representing}.
Despite the success of neural networks for paraphrase detection \citep[e.g.][]{socher2011dynamic, hu2014convolutional, yin2015convolutional}, end-to-end differentiable neural architectures failed to get close to acceptable performance for RTE due to the lack of large high-quality datasets.
An end-to-end differentiable solution to RTE is desirable, since it avoids specific assumptions about the underlying language.
In particular, there is no need for language features like part-of-speech tags or dependency parses.
Furthermore, a generic sequence-to-sequence solution allows to extend the concept of capturing entailment across any sequential data, not only natural language.

Recently, \cite{bowman2015large} published the Stanford Natural Language Inference (SNLI) corpus accompanied by a neural network with long short-term memory units \citep[LSTM,][]{hochreiter1997long}, which achieves an accuracy of $77.6\%$ for RTE on this dataset.
It is the first time a generic neural model without hand-crafted features got close to the accuracy of a simple lexicalized classifier with engineered features for RTE.
This can be explained by the high quality and size of SNLI compared to the two orders of magnitude smaller and partly synthetic datasets so far used to evaluate RTE systems.
\citeauthor{bowman2015large}'s LSTM encodes the premise and hypothesis as dense fixed-length vectors whose concatenation is subsequently used in a multi-layer perceptron (MLP) for classification.
In contrast, we are proposing an attentive neural network that is capable of reasoning over entailments of pairs of words and phrases by processing the hypothesis conditioned on the premise.

Our contributions are threefold:
(i) We present a neural model based on LSTMs that reads two sentences in one go to determine entailment, as opposed to mapping each sentence independently into a semantic space (\S\ref{sec:lstm}),
(ii) We extend this model with a neural word-by-word attention mechanism to encourage reasoning over entailments of pairs of words and phrases (\S\ref{sec:iatt}), and
(iii) We provide a detailed qualitative analysis of neural attention for RTE (\S\ref{sec:analysis}).
Our benchmark LSTM achieves an accuracy of $80.9\%$ on SNLI, outperforming a simple lexicalized classifier tailored to RTE by $2.7$ percentage points.
An extension with word-by-word neural attention surpasses this strong benchmark LSTM result by $2.6$ percentage points, setting a new state-of-the-art accuracy of $83.5\%$ for recognizing entailment on SNLI.

\section{Methods}
In this section we discuss LSTMs (\S\ref{sec:background}) and describe how they can be applied to RTE (\S\ref{sec:lstm}).
We introduce an extension of an LSTM for RTE with neural attention (\S\ref{sec:att}) and word-by-word attention (\S\ref{sec:iatt}).
Finally, we show how such attentive models can easily be used for attending both ways: over the premise conditioned on the hypothesis and over the hypothesis conditioned on the premise (\S\ref{sec:two}).

\subsection{LSTMs}
\label{sec:background}
Recurrent neural networks (RNNs) with long short-term memory (LSTM) units~\citep{hochreiter1997long} have been successfully applied to a wide range of NLP tasks, such as machine translation \citep{sutskever2014sequence}, constituency parsing \citep{vinyals2014grammar}, language modeling \citep{zaremba2014recurrent} and recently RTE \citep{bowman2015large}.
LSTMs encompass memory cells that can store information for a long period of time, as well as three types of gates that control the flow of information into and out of these cells: input gates (Eq.~\ref{input}), forget gates (Eq.~\ref{forget}) and output gates (Eq.~\ref{output}). Given an input vector $\mathbf{x}_t$ at time step $t$, the previous output $\mathbf{h}_{t-1}$ and cell state $\mathbf{c}_{t-1}$, an LSTM with hidden size $k$ computes the next output $\mathbf{h}_t$ and cell state $\mathbf{c}_t$ as

\begin{minipage}{0.35\linewidth}
  \noindent
  \begin{align}
    \mathbf{H} &= \left[{
      \begin{array}{*{20}c}
        \mathbf{x}_t \\
        \mathbf{h}_{t-1}
      \end{array} }
    \right]\\
    \mathbf{i}_t &= \sigma(\mathbf{W}^i\mathbf{H}+\mathbf{b}^i) \label{input}\\
    \mathbf{f}_t &= \sigma(\mathbf{W}^f\mathbf{H}+\mathbf{b}^f) \label{forget}
  \end{align}
\end{minipage}
\hspace{2em}
\begin{minipage}{0.53\linewidth}
  \begin{align}
    \mathbf{o}_t &= \sigma(\mathbf{W}^o\mathbf{H}+\mathbf{b}^o) \label{output}\\[.3em]
    \mathbf{c}_t &= \mathbf{f}_t \odot \mathbf{c}_{t-1} + \mathbf{i}_t \odot
    \tanh(\mathbf{W}^c\mathbf{H}+\mathbf{b}^c)\\[.3em]
    \mathbf{h}_t &= \mathbf{o}_t \odot \tanh(\mathbf{c}_t)
  \end{align}
\end{minipage}

where $\mathbf{W}^i,\mathbf{W}^f,\mathbf{W}^o,\mathbf{W}^c\in\mathbb{R}^{2k\times k}$ are trained matrices and $\mathbf{b}^i, \mathbf{b}^f, \mathbf{b}^o, \mathbf{b}^c \in \mathbb{R}^k$ trained biases that parameterize the gates and transformations of the input, $\sigma$ denotes the element-wise application of the sigmoid function and $\odot$ the element-wise multiplication of two vectors.

\subsection{Recognizing Textual Entailment with LSTMs}

\begin{figure*}[t]
\includegraphics[width=0.9\linewidth]{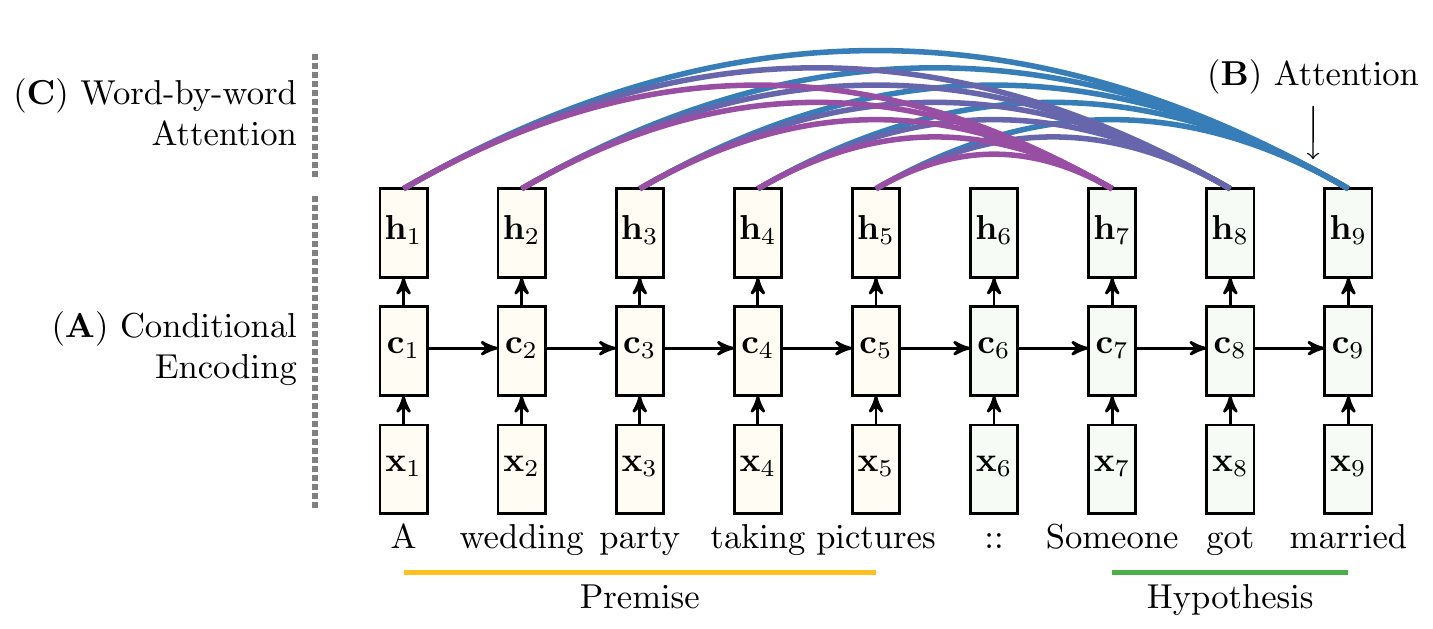}
\caption{Recognizing textual entailment using (\textbf{A}) conditional encoding via two LSTMs, one over the premise and one over the hypothesis conditioned on the representation of the premise ($\mathbf{c}_5$), (\textbf{B}) attention only based on the last output vector ($\mathbf{h}_9$) or (\textbf{C}) word-by-word attention based on all output vectors of the hypothesis ($\mathbf{h}_7$, $\mathbf{h}_8$ and $\mathbf{h}_9$).}
\label{fig:overview}
\end{figure*}

\label{sec:lstm}
LSTMs can readily be used for RTE by independently encoding the premise and hypothesis as dense vectors and taking their concatenation as input to an MLP classifier \citep{bowman2015large}. This demonstrates that LSTMs can learn semantically rich sentence representations that are suitable for determining textual entailment.

\subsubsection{Conditional Encoding}
In contrast to learning sentence representations, we are interested in neural models that read both sentences to determine entailment, thereby reasoning over entailments of pairs of words and phrases.
Figure \ref{fig:overview} shows the high-level structure of this model. The premise (left) is read by an LSTM.
A second LSTM with different parameters is reading a delimiter and the hypothesis (right), but its memory state is initialized with the last cell state of the previous LSTM ($\mathbf{c}_5$ in the example), i.e.~it is conditioned on the representation that the first LSTM built for the premise (\textbf{A}).
We use word2vec vectors \citep{mikolov2013distributed} as word representations, which we do not optimize during training.
Out-of-vocabulary words in the training set are randomly initialized by sampling values uniformly from $(-0.05,0.05)$ and optimized during training.\footnote{We found $12.1$k words in SNLI for which we could not obtain word2vec embeddings, resulting in $3.65$M tunable parameters.}
Out-of-vocabulary words encountered at inference time on the validation and test corpus are set to fixed random vectors.
By not tuning representations of words for which we have word2vec vectors, we ensure that at inference time their representation stays close to unseen similar words for which we have word2vec embeddings.
We use a linear layer to project word vectors to the dimensionality of the hidden size of the LSTM, yielding input vectors $\mathbf{x}_i$.
Finally, for classification we use a softmax layer over the output of a non-linear projection of the last output vector ($\mathbf{h}_9$ in the example) into the target space of the three classes (\textsc{Entailment}, \textsc{Neutral} or \textsc{Contradiction}), and train using the cross-entropy loss.

\subsection{Attention}
\label{sec:att}

Attentive neural networks have recently demonstrated success in a wide range of tasks ranging from handwriting synthesis \citep{graves2013generating}, digit classification \citep{mnih2014recurrent}, machine translation \citep{bahdanau2015neural}, image captioning \citep{xu2015show}, speech recognition \citep{chorowski2015attention} and sentence summarization \citep{rush2015neural}, to geometric reasoning \citep{vinyals2015pointer}.
The idea is to allow the model to attend over past output vectors (see Figure \ref{fig:overview} \textbf{B}), thereby mitigating the LSTM's cell state bottleneck.
More precisely, an LSTM with attention for RTE does not need to capture the whole semantics of the premise in its cell state.
Instead, it is sufficient to output vectors while reading the premise and accumulating a representation in the cell state that informs the second LSTM which of the output vectors of the premise it needs to attend over to determine the RTE class.

Let $\mathbf{Y} \in \mathbb{R}^{k\times L}$ be a matrix consisting of output vectors $\left[\mathbf{h}_1 \;\cdots\; \mathbf{h}_L\right]$ that the first LSTM produced when reading the $L$ words of the premise, where $k$ is a hyperparameter denoting the size of embeddings and hidden layers.
Furthermore, let $\mathbf{e}_L \in \mathbb{R}^L$ be a vector of $1$s and $\mathbf{h}_N$ be the last output vector after the premise and hypothesis were processed by the two LSTMs respectively.
The attention mechanism will produce a vector $\alpha$ of attention weights and a weighted representation $\mathbf{r}$ of the premise via
\begin{align}
  \mathbf{M} &= \tanh(\mathbf{W}^y\mathbf{Y}+\mathbf{W}^h\mathbf{h}_N\otimes \mathbf{e}_L)&\mathbf{M}&\in\mathbb{R}^{k \times L}\\
  \alpha &= \text{softmax}(\mathbf{w}^T\mathbf{M})&\alpha&\in\mathbb{R}^L\\
  \mathbf{r} &= \mathbf{Y}\alpha^T&\mathbf{r}&\in\mathbb{R}^k
\end{align}
where $\mathbf{W}^y, \mathbf{W}^h \in \mathbb{R}^{k \times k}$ are trained projection matrices, $\mathbf{w} \in \mathbb{R}^k$ is a trained parameter vector and $\mathbf{w}^T$ denotes its transpose.
Note that the outer product $\mathbf{W}^h\mathbf{h}_N\otimes \mathbf{e}_L$ is repeating the linearly transformed $\mathbf{h}_N$ as many times as there are words in the premise (i.e.~$L$ times).
Hence, the intermediate attention representation $\mathbf{m}_i$ ($i$th column vector in $\mathbf{M}$) of the $i$th word in the premise is obtained from a non-linear combination of the premise's output vector $\mathbf{h}_i$ ($i$th column vector in $\mathbf{Y}$) and the transformed $\mathbf{h}_N$.
The attention weight for the $i$th word in the premise is the result of a weighted combination (parameterized by $\mathbf{w}$) of values in $\mathbf{m}_i$.

The final sentence-pair representation is obtained from a non-linear combination of the attention-weighted representation $\mathbf{r}$ of the premise and the last output vector $\mathbf{h}_N$ using
\begin{align}
  \mathbf{h}^* & = \tanh(\mathbf{W}^p\mathbf{r}+\mathbf{W}^x\mathbf{h}_N)&\mathbf{h}^*&\in\mathbb{R}^k\label{eq:att}
\end{align}
where $\mathbf{W}^p, \mathbf{W}^x \in \mathbb{R}^{k\times k}$ are trained projection matrices.

\subsection{Word-by-word Attention}
\label{sec:iatt}

For determining whether one sentence entails another it can be a good strategy to check for entailment or contradiction of individual word- and phrase-pairs.
To encourage such behavior we employ neural word-by-word attention similar to \cite{bahdanau2015neural}, \cite{hermann2015teaching} and \cite{rush2015neural}.
The difference is that we do not use attention to generate words, but to obtain a sentence-pair encoding from fine-grained reasoning via soft-alignment of words and phrases in the premise and hypothesis.
In our case, this amounts to attending over the first LSTM's output vectors of the premise while the second LSTM processes the hypothesis one word at a time, thus generating attention weight-vectors $\alpha_t$ over all output vectors of the premise for every word $\mathbf{x}_t$ with $t\in(L+1, N)$ in the hypothesis (Figure \ref{fig:overview} \textbf{C}).
This can be modeled as follows:
\begin{align}
  \mathbf{M}_t &= \tanh(\mathbf{W}^y\mathbf{Y}+(\mathbf{W}^h\mathbf{h}_t+\mathbf{W}^r\mathbf{r}_{t-1})\otimes \mathbf{e}_L) & \mathbf{M}_t &\in\mathbb{R}^{k\times L}\label{eq:att-rec}\\
  \alpha_t &= \text{softmax}(\mathbf{w}^T\mathbf{M}_t)&\alpha_t&\in\mathbb{R}^L\\
  \mathbf{r}_t &= \mathbf{Y}\alpha^T_t + \tanh(\mathbf{W}^t\mathbf{r}_{t-1})&\mathbf{r}_t&\in\mathbb{R}^k\label{eq:att-final}
\end{align}
Note that $\mathbf{r}_t$ is dependent on the previous attention representation $\mathbf{r}_{t-1}$ to inform the model about what was attended over in the previous step (see Eq. \ref{eq:att-rec} and \ref{eq:att-final}).

As in the previous section, the final sentence-pair representation is obtained from a non-linear combination of the last attention-weighted representation of the premise (here based on the last word of the hypothesis) $\mathbf{r}_N$ and the last output vector using
\begin{align}
  \mathbf{h}^* & = \tanh(\mathbf{W}^p\mathbf{r}_N+\mathbf{W}^x\mathbf{h}_N)&\mathbf{h}^*&\in\mathbb{R}^k\label{eq:iatt}
\end{align}

\subsection{Two-way Attention}
\label{sec:two}
Inspired by bidirectional LSTMs that read a sequence and its reverse for improved encoding \citep{graves2005framewise}, we introduce two-way attention for RTE.
The idea is to use the same model (i.e.~same structure and weights) to attend over the premise conditioned on the hypothesis, as well as to attend over the hypothesis conditioned on the premise, by simply swapping the two sequences.
This produces two sentence-pair representations that we concatenate for classification.

\section{Experiments}
We conduct experiments on the Stanford Natural Language Inference corpus \cite[SNLI,][]{bowman2015large}.
This corpus is two orders of magnitude larger than other existing RTE corpora such as Sentences Involving Compositional Knowledge \citep[SICK,][]{marelli2014semeval}.
Furthermore, a large part of training examples in SICK were generated heuristically from other examples. In contrast, all sentence-pairs in SNLI stem from human annotators.
The size and quality of SNLI make it a suitable resource for training neural architectures such as the ones proposed in this paper.

We use ADAM \citep{kingma2015adam} for optimization with a first momentum coefficient of $0.9$ and a second momentum coefficient of $0.999$.\footnote{Standard configuration recommended by \citeauthor{kingma2015adam}.}
For every model we perform a small grid search over combinations of the initial learning rate [1\textsc{e}-4, 3\textsc{e}-4, 1\textsc{e}-3], dropout\footnote{As in \cite{zaremba2014recurrent}, we apply dropout only on the inputs and outputs of the network.} [0.0, 0.1, 0.2] and $\ell_2$-regularization strength [0.0, 1\textsc{e}-4, 3\textsc{e}-4, 1\textsc{e}-3].
Subsequently, we take the best configuration based on performance on the validation set, and evaluate only that configuration on the test set.

\section{Results and Discussion}

\begin{table*}
  \caption{Results on the SNLI corpus.}
  \label{tab:results}
  \centering
  \begin{tabular}{lllllll}
  	\toprule
  	Model & $k$ & $|\theta|_\text{W+M}$ & $|\theta|_\text{M}$ & Train & Dev & Test\\
  	\midrule
  	Lexicalized classifier \citep{bowman2015large} & - & - & - &  99.7 & - & 78.2\\
  	LSTM \citep{bowman2015large} & 100 & $\approx 10$M & $221$k & 84.4 & - & 77.6\\
  	\midrule
  	Conditional encoding, shared & 100 & $3.8$M & $111$k & 83.7 & 81.9 & 80.9\\
    Conditional encoding, shared & 159 & $3.9$M & $252$k & 84.4 & 83.0 & 81.4\\
    Conditional encoding & 116 & $3.9$M & $252$k & 83.5 & 82.1 & 80.9\\
  	\midrule
  	Attention & 100 & $3.9$M & $242$k & 85.4 & 83.2 & 82.3\\
    Attention, two-way & 100 & $3.9$M & $242$k & 86.5 & 83.0 & 82.4 \\
  	\midrule
  	Word-by-word attention & 100 & $3.9$M & $252$k & 85.3 & \textbf{83.7} & \textbf{83.5}\\
    Word-by-word attention, two-way & 100 & $3.9$M & $252$k & 86.6 & 83.6 & 83.2 \\
  	\bottomrule
  \end{tabular}
\end{table*}

Results on the SNLI corpus are summarized in Table \ref{tab:results}.
The total number of model parameters, including tunable word representations, is denoted by $|\theta|_\text{W+M}$ (without word representations $|\theta|_\text{M}$).
To ensure a comparable number of parameters to \citeauthor{bowman2015large}'s model that encodes the premise and hypothesis independently using one LSTM, we also run experiments for conditional encoding where the parameters of both LSTMs are shared (``Conditional encoding, shared'' with $k=100$) as opposed to using two independent LSTMs.
In addition, we compare our attentive models to two benchmark LSTMs whose hidden sizes were chosen so that they have at least as many parameters as the attentive models.
Since we are not tuning word vectors for which we have word2vec embeddings, the total number of parameters $|\theta|_\text{W+M}$ of our models is considerably smaller.
We also compare our models against the benchmark lexicalized classifier used by \citeauthor{bowman2015large}, which constructs features from the BLEU score between the premise and hypothesis, length difference, word overlap, uni- and bigrams, part-of-speech tags, as well as cross uni- and bigrams.

\paragraph{Conditional Encoding}
We found that processing the hypothesis conditioned on the premise instead of encoding each sentence independently gives an improvement of $3.3$ percentage points in accuracy over \citeauthor{bowman2015large}'s LSTM.
We argue this is due to information being able to flow from the part of the model that processes the premise to the part that processes the hypothesis.
Specifically, the model does not waste capacity on encoding the hypothesis (in fact it does not need to encode the hypothesis at all), but can read the hypothesis in a more focused way by checking words and phrases for contradictions and entailments based on the semantic representation of the premise.
One interpretation is that the LSTM is approximating a finite-state automaton for RTE \cite[cf.][]{angeli2014naturalli}.
Another difference to \citeauthor{bowman2015large}'s model is that we are using word2vec instead of GloVe for word representations and, more importantly, do not fine-tune these word embeddings.
The drop in accuracy from train to test set is less severe for our models, which suggest that fine-tuning word embeddings could be a cause of overfitting.

Our LSTM outperforms a simple lexicalized classifier by $2.7$ percentage points.
To the best of our knowledge, this is the first instance of a neural end-to-end differentiable model to achieve state-of-the-art performance on a textual entailment dataset.

\paragraph{Attention} By incorporating an attention mechanism we found a $0.9$ percentage point improvement over a single LSTM with a hidden size of $159$, and a $1.4$ percentage point increase over a benchmark model that uses two LSTMs for conditional encoding (one for the premise and one for the hypothesis conditioned on the representation of the premise).
The attention model produces output vectors summarizing contextual information of the premise that is useful to attend over later when reading the hypothesis.
Therefore, when reading the premise, the model does not have to build up a semantic representation of the whole premise, but instead a representation that helps attending over the right output vectors when processing the hypothesis.
In contrast, the output vectors of the premise are not used by the benchmark LSTMs.
Thus, these models have to build up a representation of the whole premise and carry it over through the cell state to the part that processes the hypothesis---a bottleneck that can be overcome to some degree by using attention.

\paragraph{Word-by-word Attention} Enabling the model to attend over output vectors of the premise for every word in the hypothesis yields another $1.2$ percentage point improvement compared to attending based only on the last output vector of the premise.
We argue that this is due to the model being able to check for entailment or contradiction of individual words and phrases in the hypothesis, and demonstrate this effect in the qualitative analysis below.

\paragraph{Two-way Attention}
Allowing the model to also attend over the hypothesis based on the premise does not seem to improve performance for RTE.
We suspect that this is due to entailment being an asymmetric relation.
Hence, using the same LSTM to encode the hypothesis (in one direction) and the premise (in the other direction) might lead to noise in the training signal.
This could be addressed by training different LSTMs at the cost of doubling the number of model parameters.

\subsection{Qualitative Analysis}
\label{sec:analysis}
It is instructive to analyze which output representations the model is attending over when deciding the class of an RTE example.
Note that interpretations based on attention weights have to be taken with care  since the model is not forced to solely rely on representations obtained from attention (see $\mathbf{h}_N$ in Eq.~\ref{eq:att} and \ref{eq:iatt}).
In the following we visualize and discuss the attention patterns of the presented attentive models.
For each attentive model we hand-picked examples from ten randomly drawn samples of the validation set.

\paragraph{Attention}
\begin{figure}[t!]
  \begin{subfigure}[t]{0.5\linewidth}
    \frame{
      \includegraphics[height=8.25em]{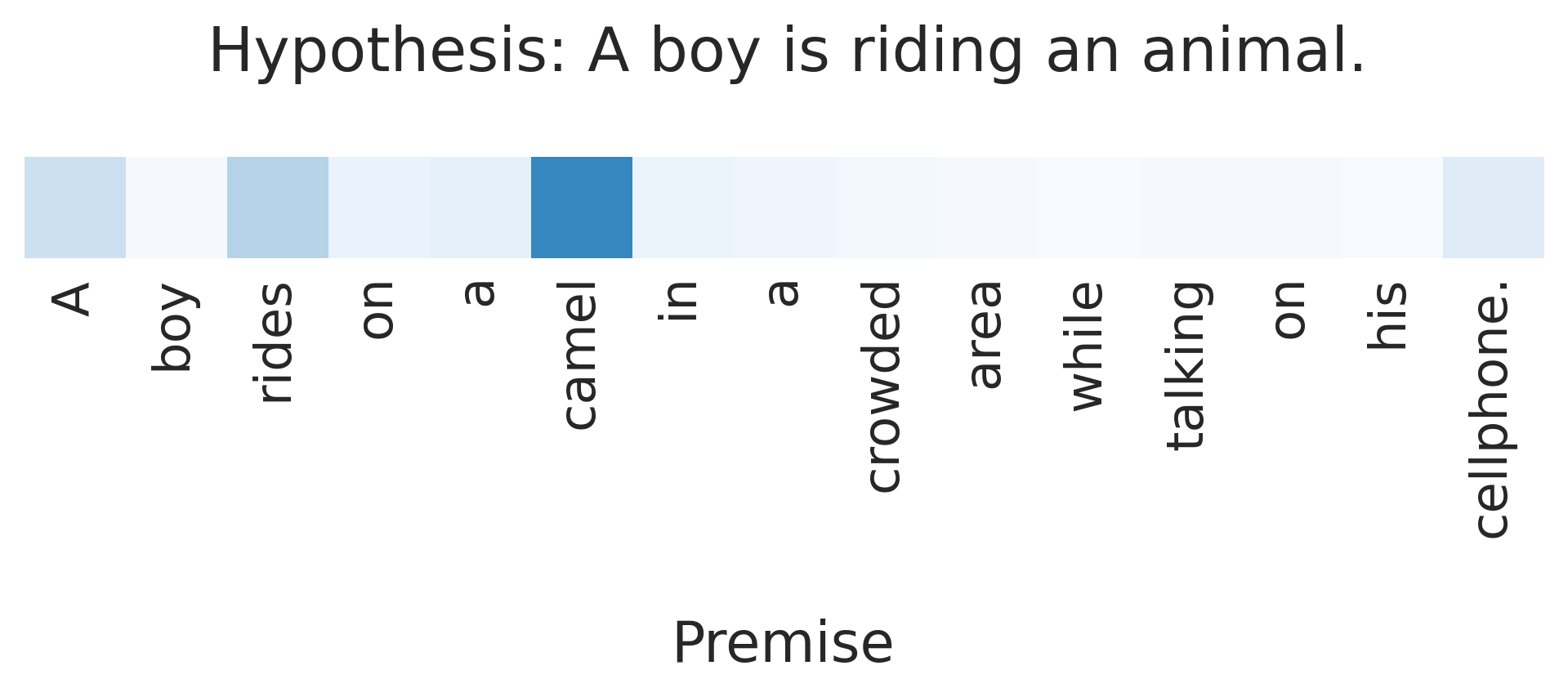}
    }
    \caption{}
    \label{fig:att:a}
  \end{subfigure}
  \begin{subfigure}[t]{0.5\linewidth}
    \hfill
    \frame{
      \includegraphics[height=8.25em]{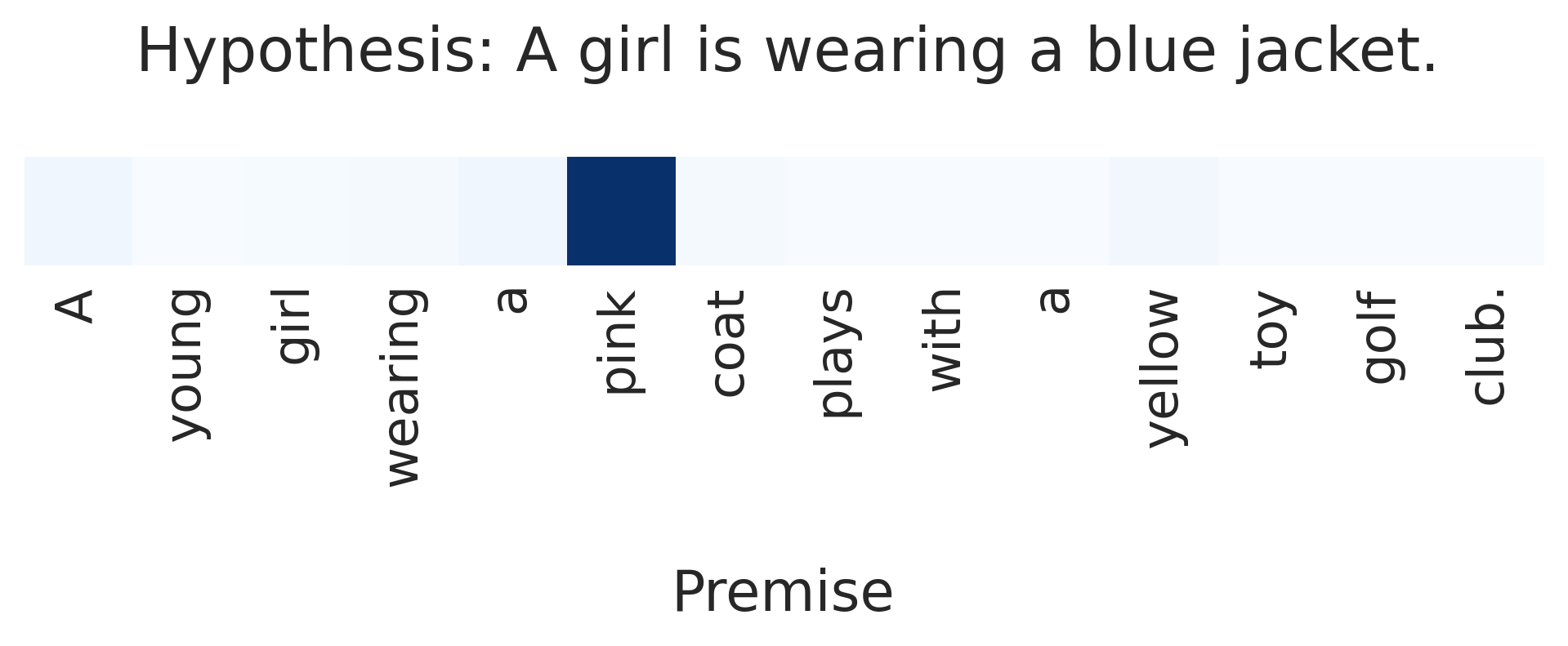}
    }
    \caption{\vspace{1.5em}}
    \label{fig:att:b}
  \end{subfigure}
  \begin{subfigure}[t]{0.47\linewidth}
    \frame{
      \includegraphics[height=8.25em]{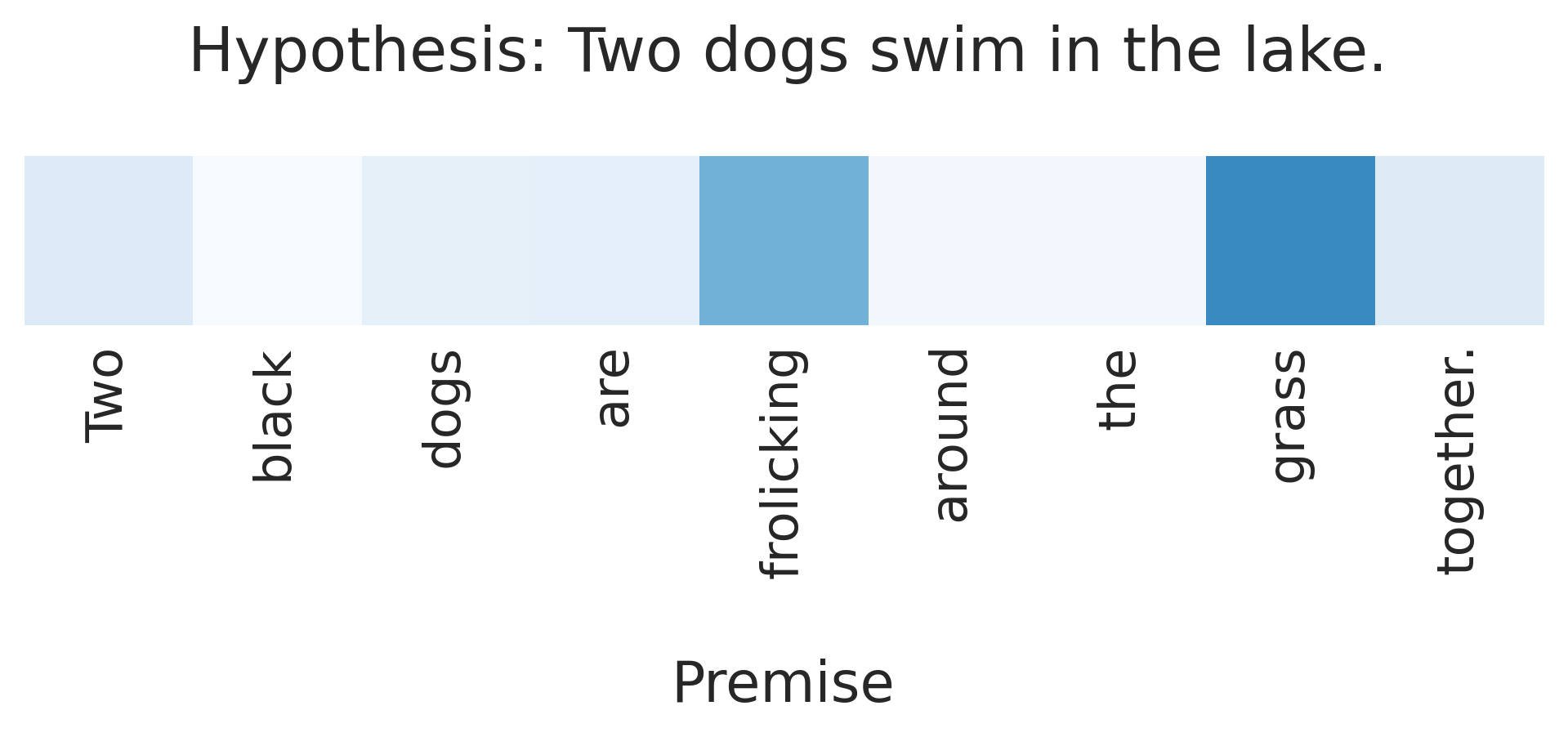}
    }
    \caption{}
    \label{fig:att:c}
  \end{subfigure}
  \begin{subfigure}[t]{0.53\linewidth}
    \hfill
    \frame{
      \includegraphics[height=8.25em]{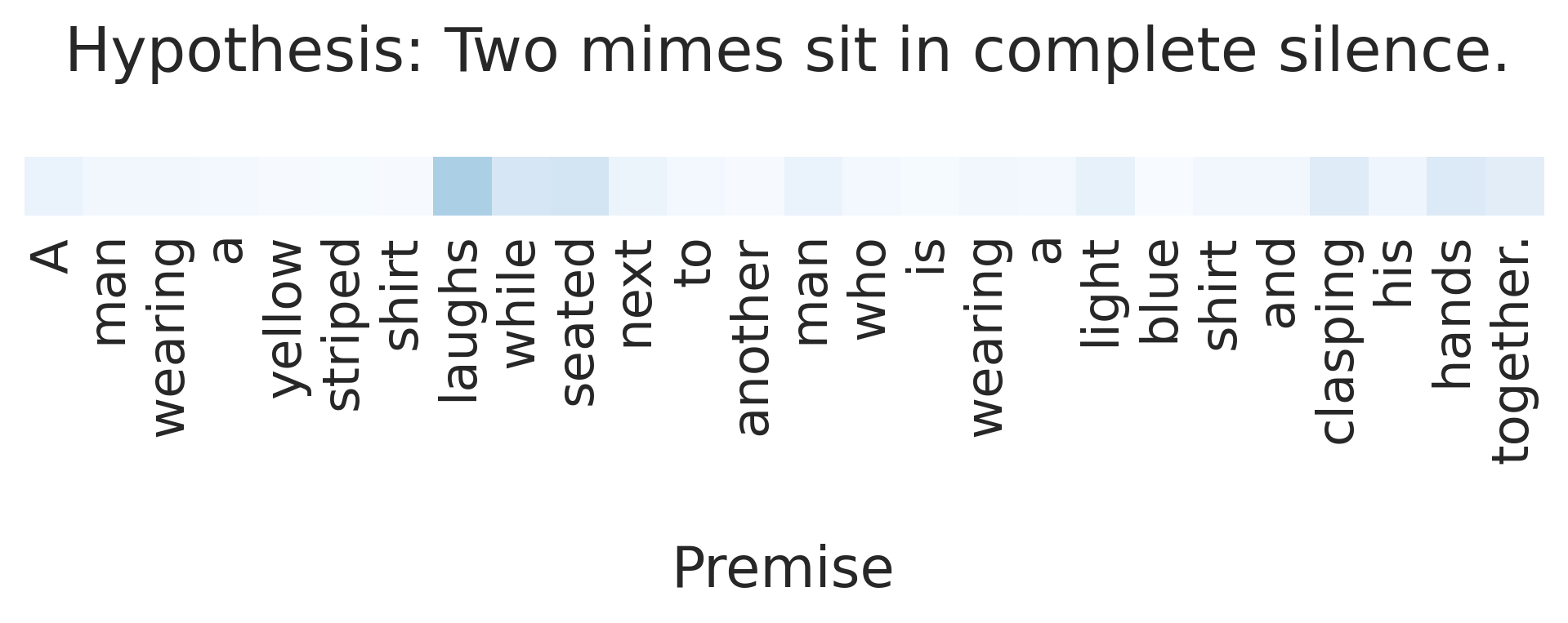}
    }
    \caption{}
    \label{fig:att:d}
  \end{subfigure}
  \caption{Attention visualizations.}
  \label{fig:att}
\end{figure}

\begin{figure}[h!]
  \begin{subfigure}[t]{0.4\linewidth}
    \frame{
      \includegraphics[height=16em]{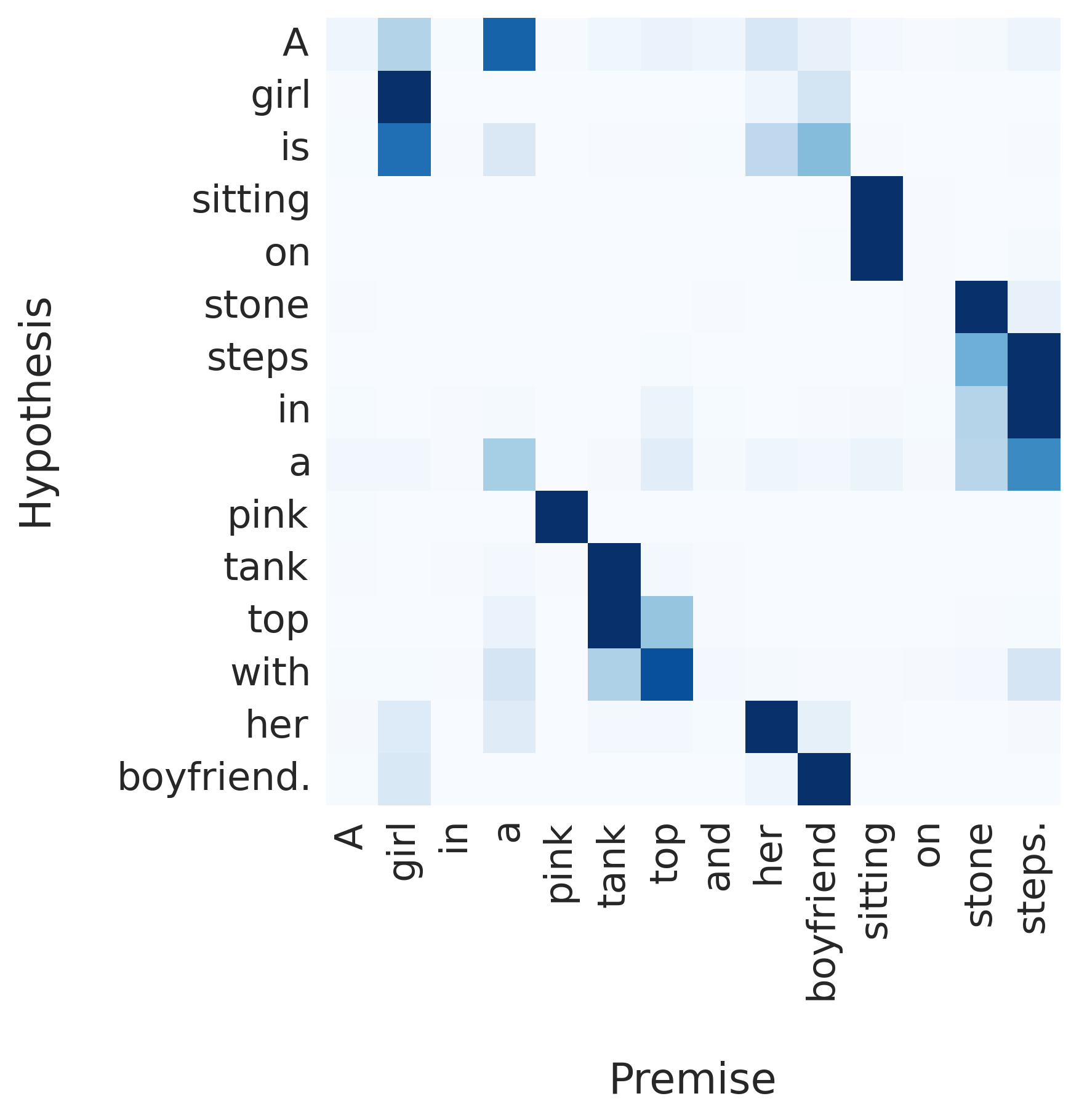}
    }
    \caption{}
    \label{fig:iatt1:a}
  \end{subfigure}
  \begin{subfigure}[t]{0.6\linewidth}
    \hfill
    \frame{
      \includegraphics[height=16em]{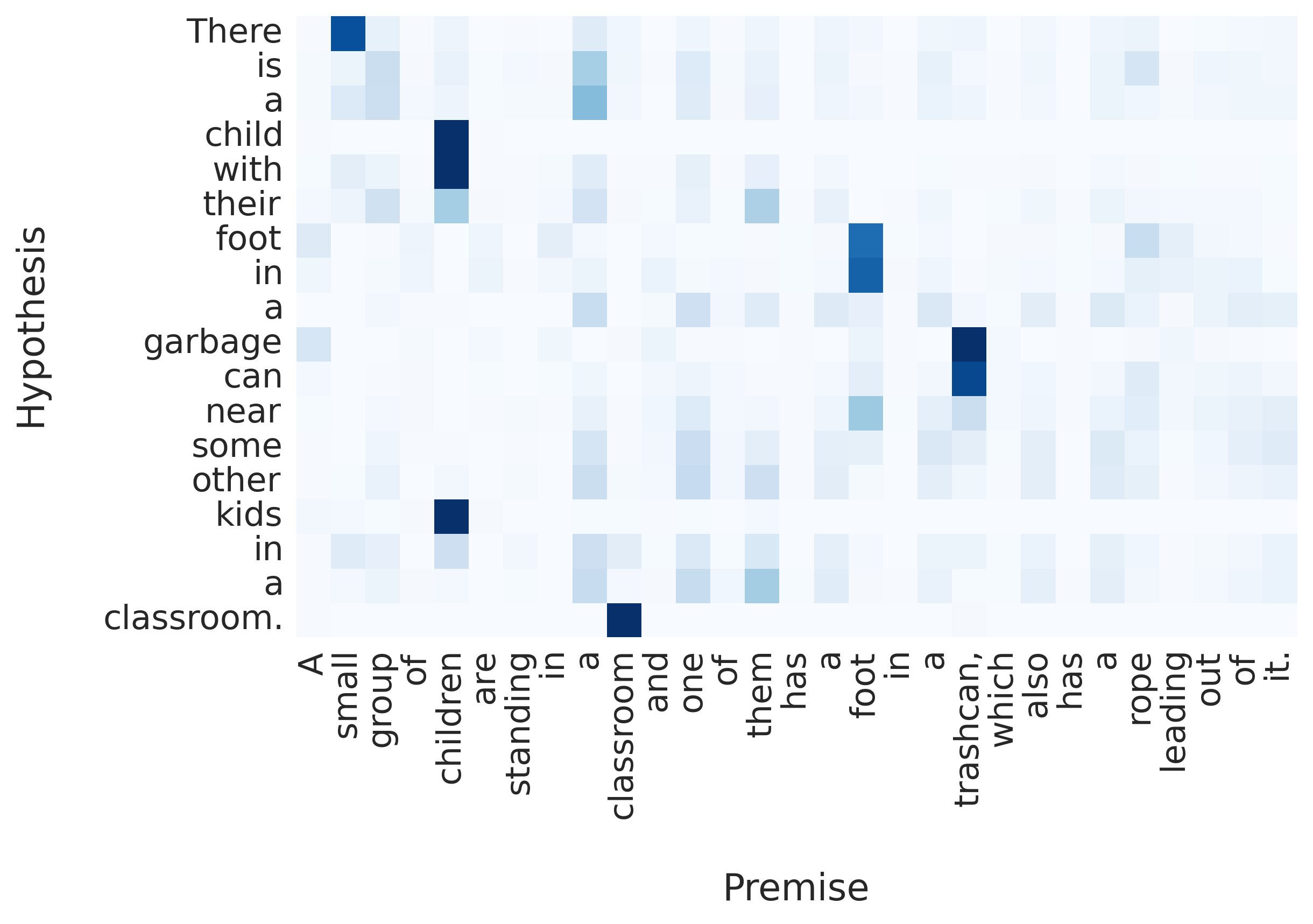}
    }
    \caption{\vspace{1.5em}}
    \label{fig:iatt1:c}
  \end{subfigure}
  \begin{subfigure}[t]{0.55\linewidth}
    \frame{
      \includegraphics[height=12.25em]{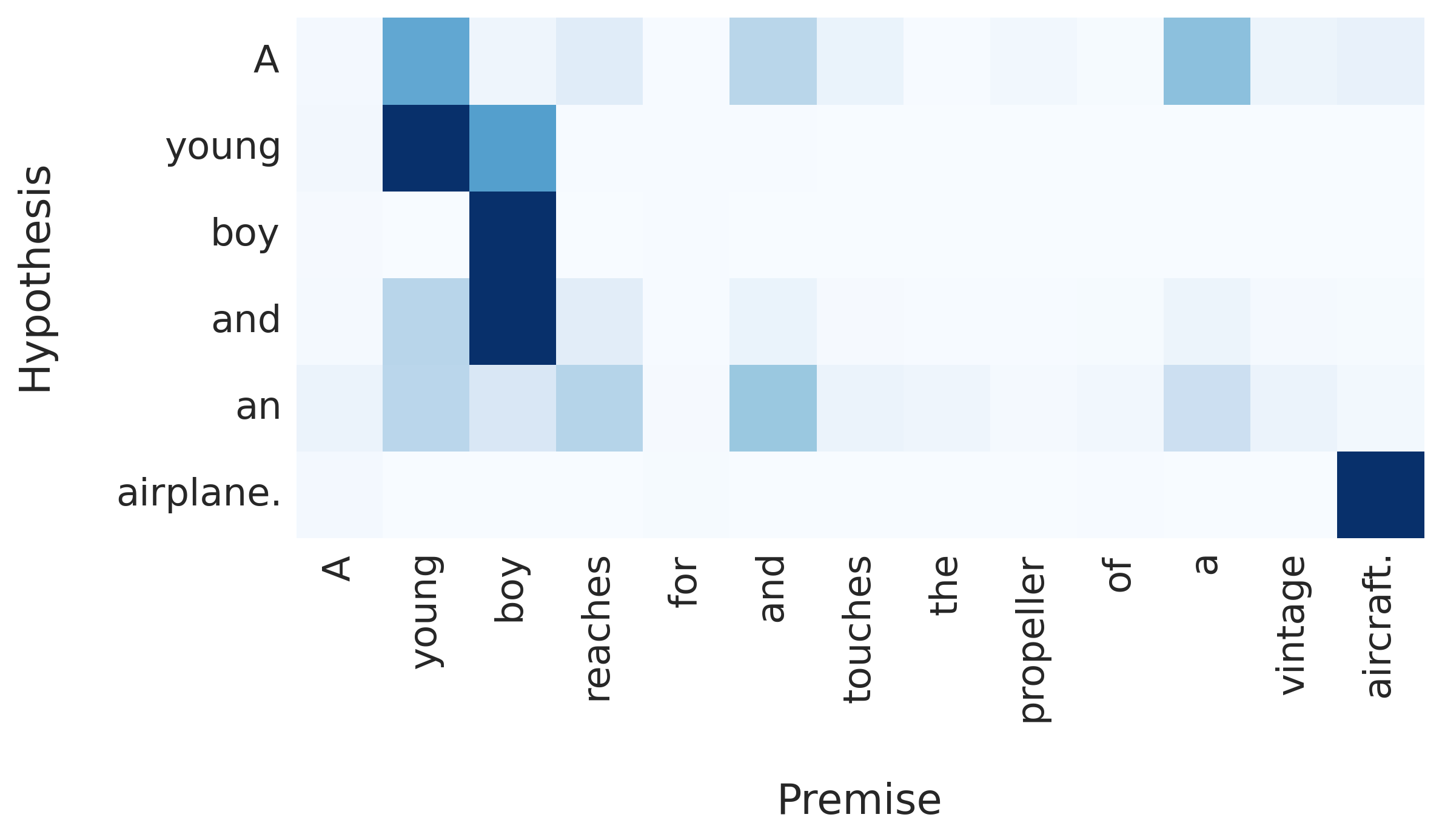}
    }
    \caption{}
    \label{fig:iatt1:b}
  \end{subfigure}
  \begin{subfigure}[t]{0.45\linewidth}
    \hfill
    \frame{
      \includegraphics[height=12.25em]{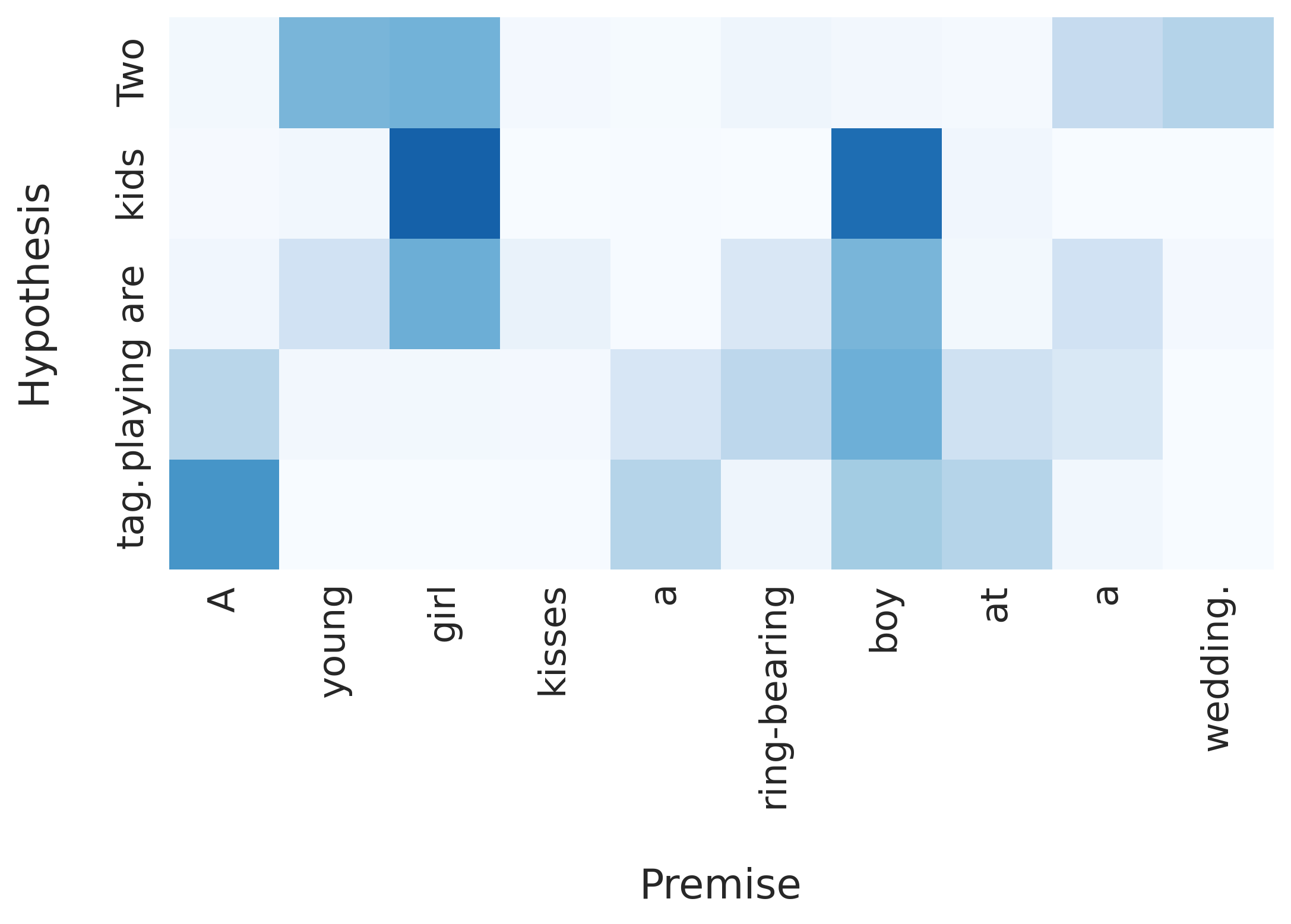}
    }
    \caption{\vspace{1.5em}}
    \label{fig:iatt2:c}
  \end{subfigure}
  \begin{subfigure}[t]{0.5\linewidth}
    \frame{
      \includegraphics[height=9em]{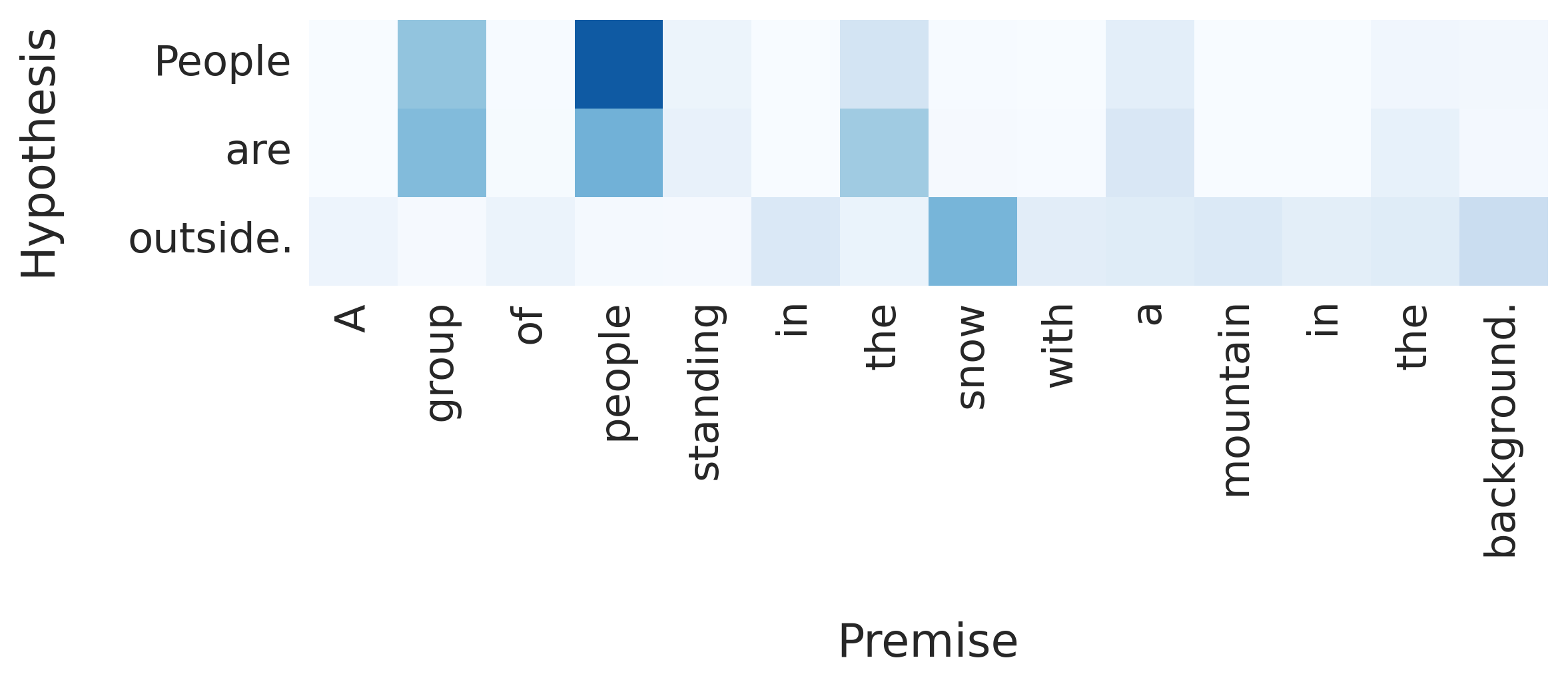}
    }
    \caption{}
    \label{fig:iatt2:a}
  \end{subfigure}
  \begin{subfigure}[t]{0.5\linewidth}
      \hfill
      \frame{
      \includegraphics[height=9em]{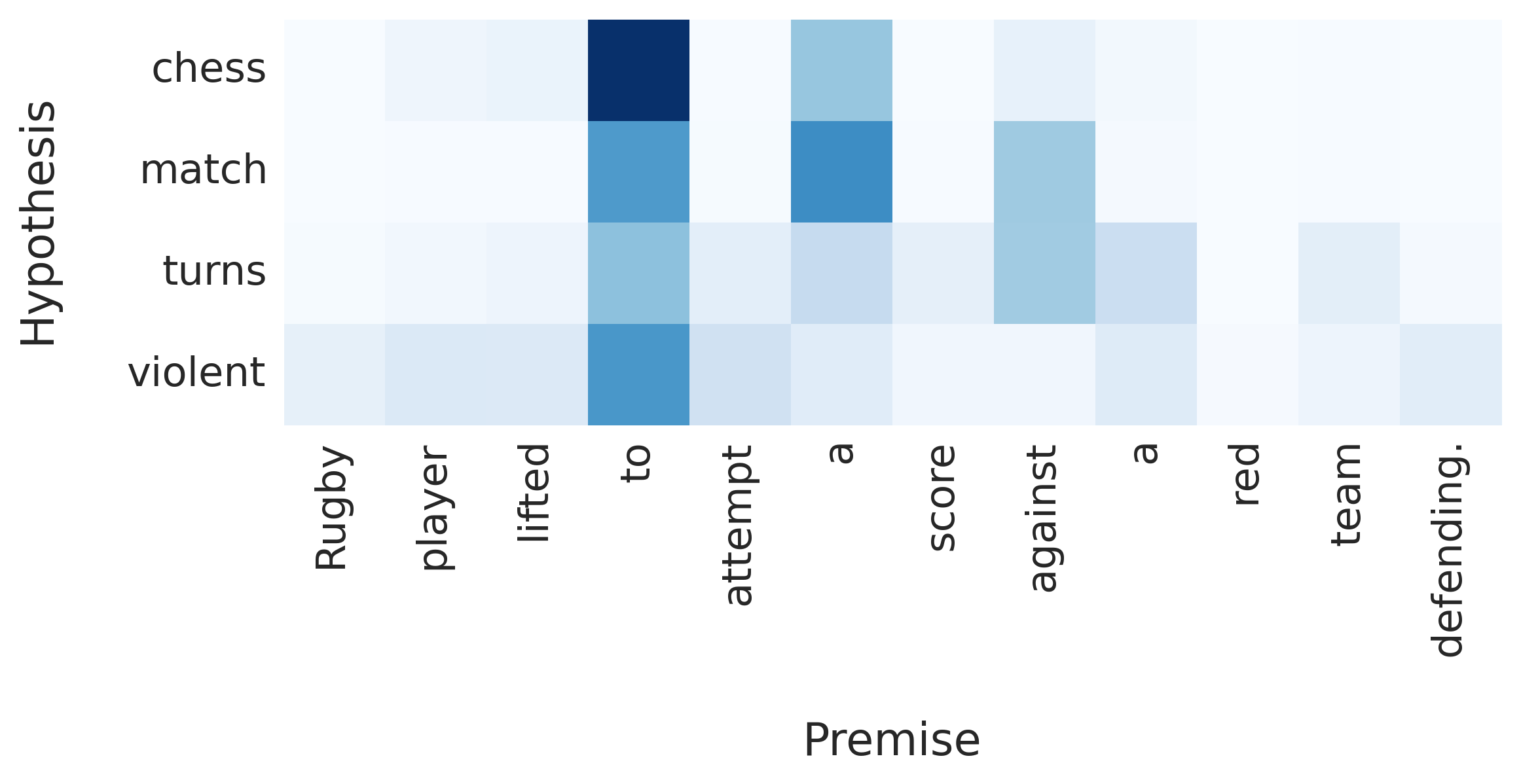}
      }
      \caption{\vspace{1.5em}}
      \label{fig:iatt3}
  \end{subfigure}
  \begin{subfigure}[t]{1.0\linewidth}
    \centering
    \frame{
      \includegraphics[height=12.75em]{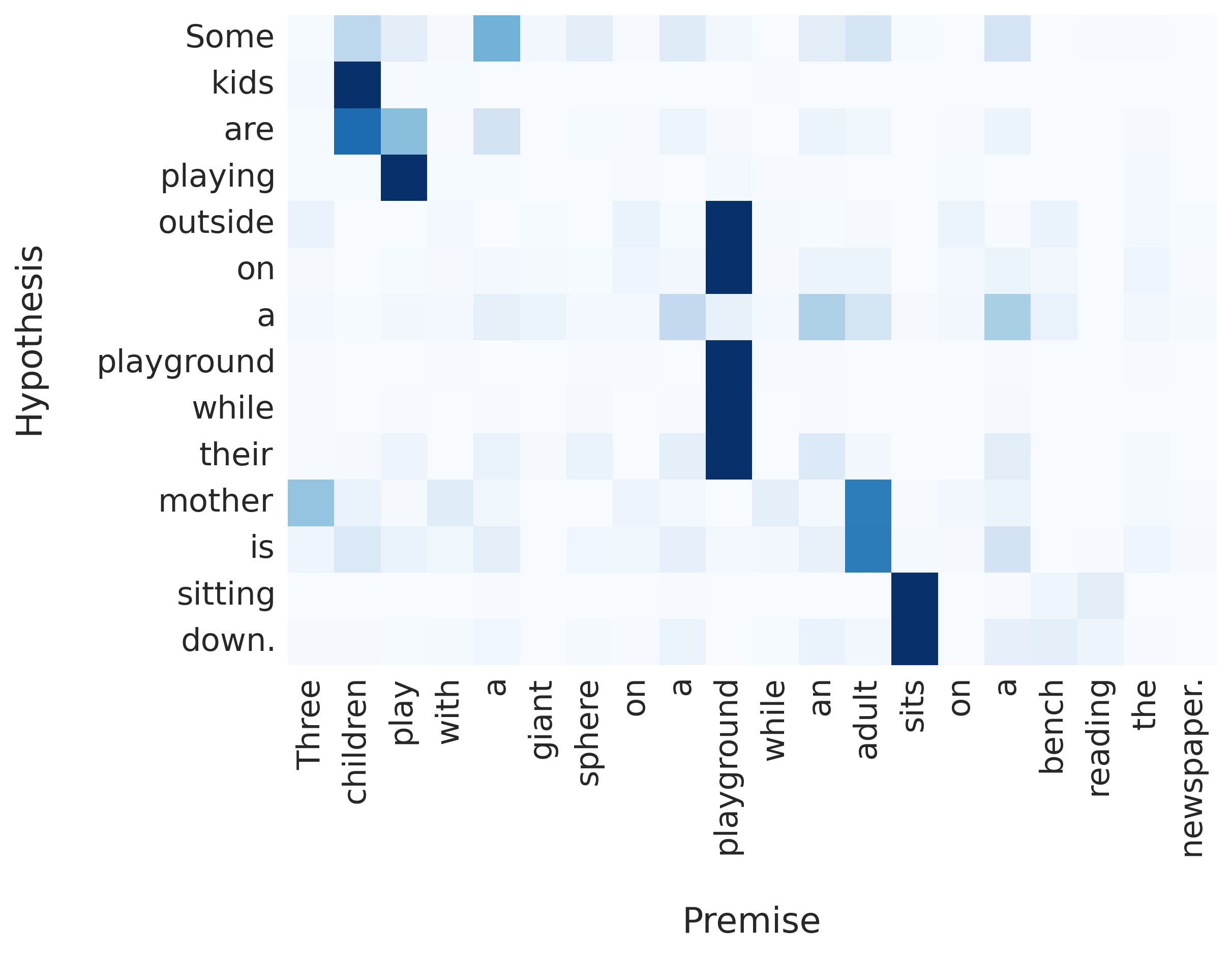}
    }
    \caption{}
    \label{fig:iatt2:b}
  \end{subfigure}
  \caption{Word-by-word attention visualizations.}
  \label{fig:iatt1}
\end{figure}

Figure \ref{fig:att} shows to what extent the attentive model focuses on contextual representations of the premise after both LSTMs processed the premise and hypothesis respectively.
Note how the model pays attention to output vectors of words that are semantically coherent with the premise (``riding'' and ``rides'', ``animal'' and ``camel'', \ref{fig:att:a}) or in contradiction, as caused by a single word (``blue'' vs. ``pink'', \ref{fig:att:b}) or multiple words (``swim'' and ``lake'' vs. ``frolicking'' and ``grass'', \ref{fig:att:c}).
Interestingly, the model shows contextual understanding by not attending over ``yellow'', the color of the toy, but ``pink'', the color of the coat.
However, for more involved examples with longer premises we found that attention is more uniformly distributed (\ref{fig:att:d}).
This suggests that conditioning attention only on the last output has limitations when multiple words need to be considered for deciding the RTE class.

\paragraph{Word-by-word Attention}
Visualizations of word-by-word attention are depicted in Figure~\ref{fig:iatt1}.
We found that word-by-word attention can easily detect if the hypothesis is simply a reordering of words in the premise (\ref{fig:iatt1:a}).
Furthermore, it is able to resolve synonyms (``airplane'' and ``aircraft'', \ref{fig:iatt1:b}) and capable of matching multi-word expressions to single words (``garbage can'' to ``trashcan'', \ref{fig:iatt1:c}).
It is also noteworthy that irrelevant parts of the premise, such as words capturing little meaning or whole uninformative relative clauses, are correctly neglected for determining entailment (``which also has a rope leading out of it'', \ref{fig:iatt1:c}).

Word-by-word attention seems to also work well when words in the premise and hypothesis are connected via deeper semantics or common-sense knowledge (``snow'' can be found ``outside'' and a ``mother'' is an ``adult'', \ref{fig:iatt2:a} and \ref{fig:iatt2:b}).
Furthermore, the model is able to resolve one-to-many relationships (``kids'' to ``boy'' and ``girl'', \ref{fig:iatt2:c})

Attention can fail, for example when the two sentences and their words are entirely unrelated (\ref{fig:iatt3}).
In such cases, the model seems to back up to attending over function words, and the sentence-pair representation is likely dominated by the last output vector ($\mathbf{h}_N$) instead of the attention-weighted representation (see Eq. \ref{eq:iatt}).

\section{Conclusion}

In this paper, we show how the state-of-the-art in recognizing textual entailment on a large, human-curated and annotated corpus, can be improved with general end-to-end differentiable models.
Our results demonstrate that LSTM recurrent neural networks that read pairs of sequences to produce a final representation from which a simple classifier predicts entailment, outperform both a neural baseline as well as a classifier with hand-engineered features.
Extending these models with attention over the premise provides further improvements to the predictive abilities of the system, resulting in a new state-of-the-art accuracy for recognizing entailment on the Stanford Natural Language Inference corpus.

The models presented here are general sequence models, requiring no appeal to Natural Language-specific processing beyond tokenization, and are therefore a suitable target for transfer learning through pre-training the recurrent systems on other corpora, and conversely, applying the models trained on this corpus to other entailment tasks.
Future work will focus on such transfer learning tasks, as well as scaling the methods presented here to larger units of text (e.g.~paragraphs and entire documents) using hierarchical attention mechanisms. Additionally, it would be worthwhile exploring how other, more structured forms of attention \citep[e.g.][]{graves2014neural,sukhbaatar2015end}, or other forms of differentiable memory \citep[e.g.][]{grefenstette2015learning,Joulin:2015:Stack}~could help improve performance on RTE over the neural models presented in this paper.
Furthermore, we aim to investigate the application of these generic models to non-natural language sequential entailment problems.

\subsubsection*{Acknowledgements}
 We thank Nando de Freitas, Samuel Bowman, Jonathan Berant, Danqi Chen, Christopher Manning, and the anonymous ICLR reviewers for their helpful comments on drafts of this paper.

{ \small
\bibliography{references}
\bibliographystyle{plainnat}
}
\end{document}